\documentclass[10pt,twocolumn,letterpaper]{article}

\usepackage[pagenumbers]{cvpr}

\definecolor{cvprblue}{rgb}{0.21,0.49,0.74}
\usepackage[pagebackref,breaklinks,colorlinks,allcolors=cvprblue]{hyperref}

\def\confName{CVPR}
\def\confYear{2027}

\title{Knowing What You Carry: Scalable Physics-Anchored\\World Models for Robotic Planning with Liquid Payloads}

\author{Yicheng Zhu$^{1}$ \quad Linfeng Tian$^{2}$ \quad Yang Chen$^{1}$ \quad Tao Li$^{2}$ \quad Zilin Bian$^{1,*}$\\
$^{1}$Rochester Institute of Technology \quad $^{2}$City University of Hong Kong\\
{\small $^{*}$Corresponding author: \tt zxbite@rit.edu}
}

\hypersetup{pdftitle={Knowing What You Carry: Scalable Physics-Anchored World Models for Robotic Planning with Liquid Payloads},pdfauthor={Yicheng Zhu, Linfeng Tian, Yang Chen, Tao Li, Zilin Bian}}

\begin{document}
\twocolumn[{%
\renewcommand\twocolumn[1][]{#1}%
\maketitle
\vspace{-1.2em}
\begin{center}
  \includegraphics[width=\linewidth]{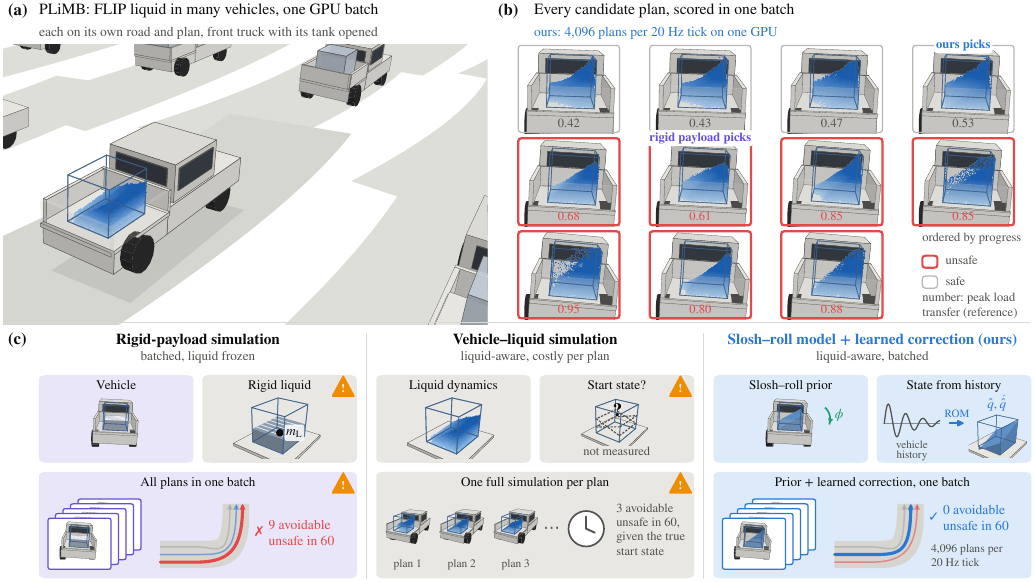}
  \captionof{figure}{Liquid payloads in a planner's world model. (a) PLiMB (Parallel Liquid--MultiBody), our GPU-parallel slosh simulator, couples FLIP liquid two-way to many vehicles in one batch, each on its own road and plan (front: tank opened). (b) One standard-tank decision: list of candidate plans with the peak load transfer ordered by progress. Red frames mark plans the reference finds unsafe. The rigid-payload planner picks an unsafe plan while ours picks the best safe one. (c) Three world models a planner can use, each with its avoidable unsafe choices in 60 standard-tank decisions at the calibrated operating point. Renders come from the simulator, started from each run's decision-time vehicle state with the liquid at rest. }
  \label{fig:teaser}
\end{center}
\vspace{0.5em}
}]
\begin{abstract}
Payload-carrying robots, such as delivery robots and autonomous trucks, must choose motions that keep them and their payload safe.
To do so, their planners predict the outcome of each candidate motion.
Modern planners compare thousands of candidates per step, so they run these predictions in parallel on a GPU, using world models such as batched simulators or learned predictors.
These world models mostly assume a rigid payload, but a liquid payload keeps moving after a maneuver, and its state is rarely measured.
Near a safety limit, small prediction errors can make an unsafe plan look safe.
Existing liquid models trade fidelity for scalability: coupled fluid simulations are accurate but too costly for every candidate, while low-order slosh models are cheap but accurate only away from resonance.
We address this trade-off for a ground vehicle with a partly filled tank near its rollover limit.
First, we build PLiMB (Parallel Liquid--MultiBody), a GPU-parallel slosh simulator that, to our knowledge, couples FLIP liquid two-way with many multibody vehicles for the first time. It simulates every candidate plan at high fidelity, providing training data and a reference for evaluation.
Second, trained on this simulator, we propose a scalable physics-anchored world model that combines a modal-slosh and roll prior, an estimate of the hidden liquid state from the vehicle's own motion history (no sensor in the tank), and a learned correction.
Without simulating the fluid, it scores thousands of candidate plans in one batched pass within a 20\,Hz control tick on one GPU.
We judge each world model by the plans it leads a planner to choose. Near the rollover limit, planners that treat the liquid as a rigid payload or use published low-order slosh models repeatedly pick unsafe plans even when a safe alternative exists. Ours never does in our standard tank and makes about half as many in a wide tank whose sloshing mode lies in the maneuver band.
There, the liquid state left by earlier driving can change the best plan, and reading it from the vehicle's history halves our model's near-limit prediction error.
\end{abstract}

\section{Introduction}
\label{sec:intro}

Payload-carrying robots, from delivery robots to autonomous trucks, must choose motions that keep themselves and their payload safe.
Their planners therefore predict the outcome of each candidate motion before acting and evaluate many candidate plans simultaneously.
Sampling-based model predictive control scores thousands of candidate trajectories per control step~\cite{williams_information_2017}, learned world models plan by rolling out imagined trajectories in batch~\cite{chua_deep_2018,hafner_dream_2020,nvidia_cosmos_2025,martinez2026independent,ijcai2026p29}, and GPU simulators step thousands of robots in parallel~\cite{makoviychuk_isaac_nodate,freeman_brax_2021,rudin_learning_2022,The_Newton_Contributors_Newton_GPU-accelerated_physics_2025}, including whole traffic scenes with physics-based vehicles~\cite{zhu_scenefactory_2026,zhu_scenefactory3d_2026}.
Most of these world models assume that the robot is a set of rigid bodies and that its payload moves rigidly with it.

However, this assumption fails for liquid payloads, which are common: manipulators and serving robots move open containers~\cite{moriello_manipulating_2018,muchacho_solution_2022,choi_suppressing_2024}, spacecraft carry propellant~\cite{abramson_dynamic_1966,huo_data-driven_2025}, and road and site vehicles carry fuel and water~\cite{kolaei_range_2014,liu_safe_2021}.
In each case, the liquid shifts when the carrier accelerates and keeps moving after the maneuver ends, so the forces on the carrier depend on how it moved seconds earlier~\cite{kolaei_range_2014,qi_observing_2023}.
Near a safety boundary (a spill, a tip-over, a rollover), small errors in the predicted load can change which plans the planner judges safe.

Existing works that account for the liquid usually trade fidelity for scalability.
High-fidelity simulators that couple the liquid to the carrier, such as Chrono~\cite{tasora_chrono_2016} and computational fluid dynamics or particle methods, are accurate but costly, because every candidate plan needs its own coupled simulation, even when batched.%
Low-order slosh models, an equivalent pendulum or a spring--mass system per sloshing mode~\cite{abramson_dynamic_1966,dodge_new_nodate}, are fast and interpretable, and recent controllers embed them in model predictive control~\cite{kolaei_coupled_2017,qi_anti-rollover_2025,liang_event-triggered_2025}. 
Yet, they are accurate only away from resonance, when the liquid's sloshing frequency lies outside the frequencies the maneuvers excite~\cite{kolaei_range_2014}.
They also need the liquid's current state, which is hidden. The work closest to ours, by Qi \etal, measures it with a free-surface sensor in the tank~\cite{qi_observing_2023,qi_anti-rollover_2025}, which production vehicles lack.
Learned world models, a third option, are fast in batch, but it is rarely clear which physical quantities, such as the hidden liquid state, they must represent, and lower prediction error does not guarantee better decisions~\cite{lambert_objective_2021,donti_task-based_2019,dauner_parting_2023,wang_not_2026}.
As a result, no existing world model is liquid-aware, 
accurate whether or not the maneuvers excite the liquid near resonance,
and cheap enough to score every candidate plan (\cref{fig:teaser}).
We ask \emph{what a planning world model needs to know about the liquid} when it must score thousands of plans, and study it on a ground vehicle carrying a partly filled tank near its rollover limit.

We first build PLiMB (Parallel Liquid--MultiBody), a GPU-parallel slosh simulator that, to our knowledge, for the first time, couples FLIP liquid~\cite{zhu_animating_nodate} two-way to many multibody vehicles at once, each on its own road and plan. 
As the high-fidelity side of the trade-off, it simulates the outcome of every candidate plan and has produced more than 5,600 coupled FLIP runs and about 32,000 runs with a reduced-order liquid (\cref{tab:supp-runs}).
Like any coupled fluid--multibody simulation, it is too slow to run inside a control loop, so we use it to train and evaluate a scalable physics-anchored world model (\cref{fig:model}).
A fast vehicle prior with a nonlinear slosh model rolls each plan forward, a causal estimate of the liquid's modal state from the vehicle's own logged history supplies the hidden state, and a learned correction fixes the prior's errors.
Everything after the state estimate is independent per plan, so the model scores all candidates in one batched GPU pass without simulating the fluid or the multibody vehicle. We judge every world model both by its prediction error and by the plans it leads a fixed planner to choose.

This makes the world model scalable. On one GPU, it scores 4,096 candidate plans within one 20\,Hz control tick, 93$\times$ the throughput of a planner that simulates a reduced-order liquid with the multibody vehicle, and batching does not change its safety decisions.

We count \emph{avoidable unsafe choices} in decisions near the rollover limit, that is, picks of a plan the simulator finds unsafe when a safe plan was available.
Treating the liquid as a rigid payload is not enough. Even with the multibody vehicle simulated and its safety margin calibrated, such a planner still makes 9 avoidable unsafe choices in 60 decisions in our standard tank.
Ours makes none, comparable to simulating a reduced-order liquid with the vehicle at far higher cost, while two published low-order slosh models adapted to the same vehicle, a pendulum after Qi \etal~\cite{qi_anti-rollover_2025} and a Kolaei-type linear multimodal model~\cite{kolaei_coupled_2017}, make 22 and 14.
In a wide tank whose sloshing mode lies in the maneuver band, our model makes about half as many as these two.
There, with the vehicle state held identical, the liquid state left by a preceding maneuver changes the best plan in most paired tests, and without reading this state from the vehicle's history, our model's near-limit prediction error doubles.

In summary, our contributions are:
\begin{itemize}
\item PLiMB, to our knowledge the first GPU-parallel slosh simulator that couples FLIP liquid two-way to many multibody vehicles at once.
\item a scalable physics-anchored world model that estimates the hidden liquid state from the vehicle's own signals and scores thousands of candidate plans per control tick on one GPU.
\item decision-level experiments showing that a rigid-payload world model leads to avoidable unsafe choices, that ours makes fewer than rigid-payload and published slosh-model planners, and that in a wide tank the liquid state left by earlier driving can change the best plan.
\end{itemize}

\section{Related Work}
\label{sec:related}

\subsection{Liquid Slosh in Vehicle Dynamics}
\label{sec:rel-slosh}
A liquid payload lowers a tank vehicle's rollover threshold relative to an equivalent rigid load, and dynamic slosh lowers it further than quasi-static load shift predicts~\cite{rakheja_research_assistant_professor_roll_1988,kolaei_range_2014}.
The standard engineering models replace the liquid by equivalent pendulums or spring--mass systems, one per sloshing mode~\cite{abramson_dynamic_1966,dodge_new_nodate}.
Derived from potential flow, they reproduce slosh loads well away from resonance and degrade near it~\cite{kolaei_range_2014}, and calibrated pendulums also accumulate phase error in free sloshing~\cite{qi_observing_2023}. Kolaei \etal couple a linear multimodal version to a full vehicle co-simulation~\cite{kolaei_coupled_2017}.
Qi \etal show that the slosh state is hidden and depends on the phase of consecutive maneuvers, and estimate it with a simulated free-surface sensor~\cite{qi_observing_2023}. They later embed a CFD-calibrated pendulum in an anti-rollover MPC~\cite{qi_anti-rollover_2025}.
Related controllers use a pendulum-based parameter-varying model in a tube MPC~\cite{liang_event-triggered_2025} or learn a rollover governor for a pendulum-slosh plant~\cite{liu_safe_2021}.
In this line the liquid model is fixed in advance and the controller is what is studied. In the work closest to ours, the liquid state comes from a free-surface sensor in the tank~\cite{qi_observing_2023,qi_anti-rollover_2025}, which production vehicles lack. We estimate that state from the vehicle's own signals, and compare liquid models by the plans they lead a fixed planner to choose.

\subsection{Physics-Informed Learned Dynamics}
Augmenting an incomplete physical model with a learned correction is well established~\cite{yin_augmenting_2021,rackauckas_universal_2021,kennedy_bayesian_2001}, and such priors are known to make learned dynamics more data-efficient~\cite{gupta_structured_2020,djeumou_neural_2022}.
In vehicle control, a learned residual over a nominal model improves MPC~\cite{kabzan_learning-based_2019}.
For liquids, learned surrogates predict slosh loads or free surfaces under prescribed excitation~\cite{huo_data-driven_2025,huang_deep_2026,olucha_data-driven_2026}, hidden liquid state has been reconstructed from free-surface history~\cite{moya_physics_2022}, and concurrent work uses a learned latent disturbance to control a liquid-carrying robot~\cite{kim_statistical_2026}.
Learned particle simulators of fluids~\cite{sanchez-gonzalez_learning_2020,ummenhofer_lagrangian_2020,toshev_lagrangebench_2023} are evaluated mainly by prediction error, and their transfer can be fragile~\cite{klimesch_simulating_2022}, while robots that transport liquid usually plan from a liquid at rest~\cite{moriello_manipulating_2018,muchacho_solution_2022,hynninen_emergency_2026}.
These works predict under prescribed excitation or from a liquid at rest. In our setting, the liquid already carries state from earlier driving, which the model must infer from the vehicle's history.

\subsection{GPU-Parallel Simulation and Planning}
The batched simulators~\cite{nvidia_isaac_2025} and simulation frameworks~\cite{Macklin_Warp_A_High-performance_2022}, sampling-based planners and world models of \cref{sec:intro} treat the payload as rigid or absent.
Fluids run on GPUs for graphics and learned simulation~\cite{brackbill_flip_1986,zhu_animating_nodate,sanchez-gonzalez_learning_2020}, and a concurrent multi-material engine couples SPH liquid two-way to robot arms in many parallel worlds on one GPU~\cite{gupta_alloy_2026}. To our knowledge, liquid has not been coupled to many vehicles at once for planning, and the liquid-aware controllers of \cref{sec:rel-slosh} optimize a single trajectory per step as published.
PLiMB couples FLIP two-way to many multibody vehicles in one batch (comparison in \cref{sec:supp-position}), and our world model scores candidate plans with the liquid in one batch.
\subsection{Decision-Aware Evaluation of Models}
In model-based reinforcement learning, likelihood correlates weakly with return~\cite{lambert_objective_2021}, and decision-aware objectives can control better with less accurate models~\cite{farahmand_value-aware_nodate,nikishin_control-oriented_2021}.
In decision-focused learning, the lowest-error forecaster can carry a higher task cost~\cite{donti_task-based_2019,elmachtoub_smart_2020}, and in driving, open-loop forecasting scores relate poorly to closed-loop planning~\cite{dauner_parting_2023,dauner_navsim_2024}.
Concurrent studies find that decision-relevant error predicts planning quality far better than total error~\cite{wang_not_2026}, an error shared across a candidate pool can dominate prediction error without changing rankings~\cite{min_when_2026}, and history-dependent state can change the best action in excavation~\cite{lin_pilebelief_2026}.
We apply this view to a hidden physical state, ablating the liquid state and the physics prior and scoring both prediction and decisions against the same reference simulator.

\section{Method}
\label{sec:method}

\subsection{Problem Formulation}
\label{sec:problem}
We consider a planner that, at a decision point, chooses one plan from a finite set of candidates.
A decision group $g$ consists of a start state, made of the vehicle state $x_g$ and a liquid state $\xi_g$ that is not observed, together with candidate plans $\Pi_g=\{\pi_1,\dots,\pi_{n_g}\}$, with $n_g$ between 8 and 16.
Each plan specifies a path and a speed profile over the next $T=5$\,s and is executed by a fixed tracking controller.
Executed in the reference simulator, plan $\pi$ yields a peak load-transfer ratio $\ell(\pi)=\max_t|\mathrm{LTR}(t)|$, with $\mathrm{LTR}=(F_\mathrm{L}-F_\mathrm{R})/(F_\mathrm{L}+F_\mathrm{R})$ computed from the left and right wheel loads, and a peak lateral tracking error $e(\pi)$.
The plan is unsafe, $u(\pi)=1$, if $\ell(\pi)\ge\tau$ or $e(\pi)\ge\bar e$, with $\tau=0.6$ and $\bar e=1.5$\,m, or if the run ends early because the vehicle rolls over or leaves the road.

\begin{figure}[t]
  \centering
  \includegraphics[width=\linewidth]{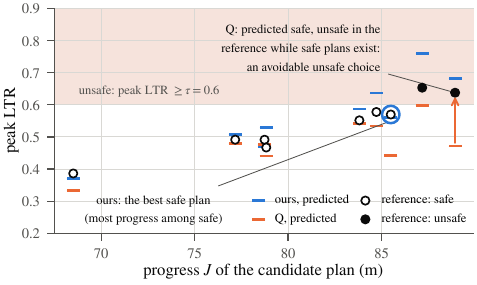}
  \caption{One standard-tank decision group at the calibrated operating point: each candidate's reference peak LTR against its progress, with the predictions of our model and of Q. Q picks the plan with the most progress among those it predicts safe, which is an avoidable unsafe choice. Ours picks the best safe plan.}
  \label{fig:setup}
\end{figure}

A planner is defined by a predictor $f$ that maps the observed history $h_g$ and a plan to predicted peaks $(\hat\ell,\hat e)$.
All planners use the same selection rule,
\begin{equation}
\pi^\ast_g=\arg\max_{\pi\in\Pi_g}\;J(\pi)\quad\text{s.t.}\quad \hat\ell(\pi)-b<\tau,\;\; \hat e(\pi)<\bar e,
\label{eq:select}
\end{equation}
where $J$ is the plan's nominal progress and $b$ is an optional bias correction (\cref{sec:setup}).
A choice is an \emph{avoidable unsafe choice} if $u(\pi^\ast_g)=1$ while some $\pi\in\Pi_g$ has $u(\pi)=0$ (\cref{fig:setup}).
The number of such groups is our primary outcome. We also report how often a planner picks the reference's best safe plan.
Since planners differ only in $f$, this outcome measures what each predictor's errors cost in decisions.

\subsection{PLiMB: GPU-Parallel Coupled Slosh Simulation}
\label{sec:sim}
PLiMB, our reference simulator, couples a GPU FLIP liquid~\cite{zhu_animating_nodate} two-way to a multibody vehicle simulated in Newton~\cite{The_Newton_Contributors_Newton_GPU-accelerated_physics_2025}.
Every vehicle carries its own tank and liquid grid, and many independent vehicles advance together in one batch on one GPU (256 in our timing benchmark, \cref{sec:scale}), each on its own road and plan, so we can label thousands of candidate plans and replay one start state under many plans.
The same vehicles can carry the liquid as a reduced-order model (ROM, \cref{sec:model}) or as a rigid payload in place of FLIP, which gives cheaper labels and the simulation-based baselines of \cref{sec:setup}.
The vehicle steps at 240\,Hz with one liquid step per vehicle step. The liquid uses a 2.5\,cm grid, a FLIP/PIC blend of 0.98, a conjugate-gradient pressure projection (50 iterations) with a 10-iteration volume correction, and a ghost-fluid free surface. The slosh wrench is exchanged with the vehicle each step with a one-step kinematic lag.
In a separate pre-registered study (\cref{sec:supp-validation}), FLIP at these settings matches a 2D volume-of-fluid solver on slosh force and roll moment, moving the static LTR by at most 0.035 against a median decision margin of 0.35. Against SPHERIC experiments, it over-predicts the roll of a freely rolling tank by 37\,\% and, like the CFD solver, the wall-pressure impulse.
The vehicle is a pickup of about 1.9\,t with a 1.65\,m track whose bed carries a rectangular $1.2\times1.0\times1.0$\,m tank filled to 40\,\% with water.
The liquid is 480\,kg, 20\,\% of the laden mass, and its first lateral sloshing mode is at 0.81\,Hz, above the 0.2--0.6\,Hz band that typical maneuvers excite.
We also use a wide tank of $1.75\times1.5\times1.0$\,m at the same fill, holding 1,050\,kg (34\,\%), whose lateral mode at 0.60\,Hz falls inside that band. This is where we expect a hidden liquid state to matter most.

\subsection{Response Model}
\label{sec:model}

\begin{figure*}[t]
  \centering
  \includegraphics[width=\linewidth]{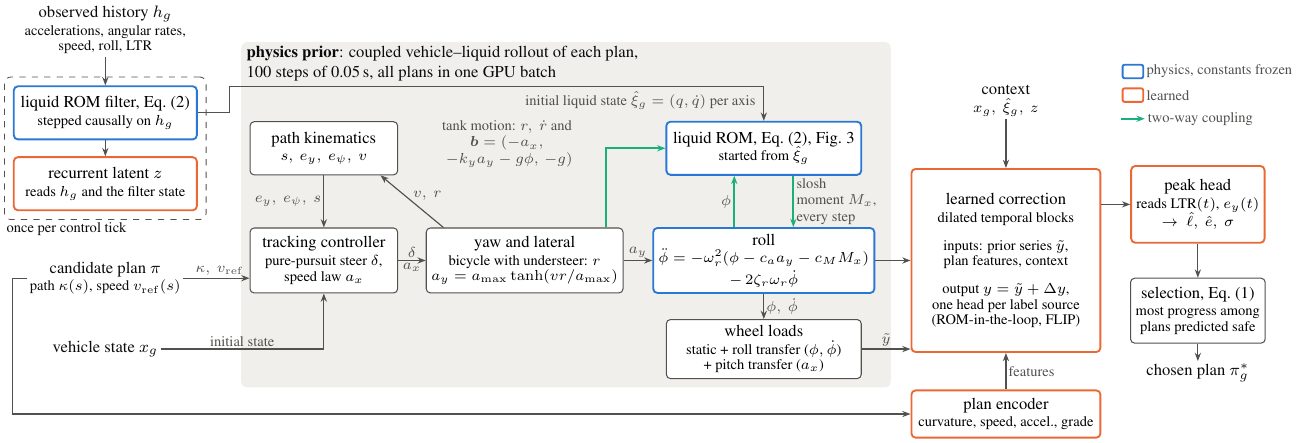}
  \caption{The physics-anchored response model. Once per tick, the liquid ROM (\cref{fig:rom}) is stepped on the vehicle's on-board signals to estimate the hidden liquid state $\hat\xi_g$, and a recurrent latent $z$ summarizes the same history. For all candidate plans in one GPU batch, the physics prior rolls vehicle and liquid forward together, and a second ROM copy, started from $\hat\xi_g$, feeds its roll moment $M_x$ back into roll (green). The fitted constants are $c_a$, $c_M$, $\omega_r$, $\zeta_r$, $a_{\max}$ and $k_y$. A learned correction, one head per label source, refines the prior's response series, and a peak head predicts $\hat\ell$ and $\hat e$ (and an uncertainty $\sigma$ that \cref{eq:select} does not use).}
  \label{fig:model}
\end{figure*}

\begin{figure}[t]
  \centering
  \includegraphics[width=\linewidth]{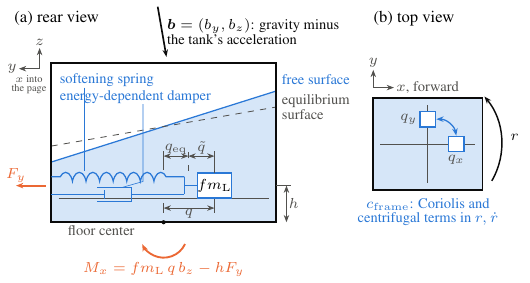}
  \caption{The liquid ROM of \cref{eq:rom}, one mode per horizontal axis. (a) Rear view: the sloshing mass $f m_\mathrm{L}$ at height $h$ is displaced by $q=q_\mathrm{eq}+\tilde q$ (its equilibrium under $\boldsymbol b$ plus the wave, drawn for $b_y<0$) and held by a softening spring and an energy-dependent damper. It loads the vehicle with $F_y$ and $M_x$ (arrows: positive directions). (b) Top view: the longitudinal mode has the same form, coupled to the lateral one through $c_\mathrm{frame}$.}
  \label{fig:rom}
\end{figure}

Our predictor $f$ maps the observed history and a batch of candidate plans to each plan's response over $T$, without any fluid or multibody simulation at inference (\cref{fig:model}).
It is built around a reduced-order model (ROM) of the liquid with one sloshing mode per horizontal axis (\cref{fig:rom}).
Each mode is an equivalent mass $f m_\mathrm{L}$, a fraction $f$ of the liquid mass $m_\mathrm{L}$, that slides at height $h$ above the tank floor, and the remaining $(1-f)m_\mathrm{L}$ moves rigidly with the tank.
For each axis, with the mass's displacement $q$ from the tank center, the ROM follows

\begin{equation}
\ddot q = - \omega^2 g^\ast\,(\tilde q + k\,\tilde q^{3}) - 2\zeta\omega\big(1+(E/E_c)^{n}\big)\,\dot q + c_\mathrm{frame},
\label{eq:rom}
\end{equation}
where $\tilde q=q-b/(\omega^2g^\ast)$ is the wave measured from the instantaneous equilibrium, with $b$ the component along the axis of $\boldsymbol b$, gravity minus the tank's acceleration in the tank frame. Here $g^\ast=1+c_h(-b_z/g-1)$ is a dimensionless heave factor, so that $\omega$ is the mode's natural frequency. The cubic term softens the mode with amplitude, damping rises with the wave energy $E=\tilde q^2+(\dot q/\omega)^2$ above a scale $E_c$, and $c_\mathrm{frame}$ collects the rotating-frame terms, through which yaw couples the two axes without doing work.
The mode's motion loads the tank with the lateral force $F_y=-f m_\mathrm{L}\ddot q$ plus rotating-frame terms, and its offset under the vertical component $b_z$ adds a roll moment about the tank floor, $M_x=f m_\mathrm{L}\,q\,b_z-hF_y$.
In the standard tank, the 17 constants of \cref{eq:rom} are identified once from FLIP and then frozen.

The ROM serves two roles.
First, stepped causally on the tank's acceleration, angular rate and angular acceleration as logged by the simulator (without sensor noise), it produces an estimate $\hat\xi_g=(q,\dot q)$ per axis of the liquid state at the decision point, without access to the true fluid.
A small recurrent latent, which also reads speed, roll and load transfer, complements it, but our ablations find that it adds little.
Second, it is part of a physics prior, a coupled vehicle--liquid rollout over each plan at 20\,Hz that starts from $x_g$ and $\hat\xi_g$ (\cref{fig:model}).
At each step, the controller's own tracking law sets steering and longitudinal acceleration $a_x$. A bicycle model with understeer gives the yaw rate $r$ and a saturating lateral acceleration $a_y$, and a second-order roll model is driven by $a_y$ and by the slosh moment $M_x$.
The vehicle's motion, $\boldsymbol b=(-a_x,\,-k_ya_y-g\phi,\,-g)$ in the tank frame (with a fitted gain $k_y$, and $g^\ast=1$) together with $r$ and $\dot r$, forces the ROM, whose moment enters the roll equation at the next step, so the liquid and the vehicle are coupled both ways.
The four wheel loads, and hence the LTR, follow from roll, roll rate and $a_x$.
The prior has 21 constants, fitted to simulation runs with the ROM in the loop. For deployment in the standard tank, it can be compiled into a single GPU kernel (\cref{sec:scale}).

The learned correction then refines the prior.
A plan encoder reads each plan on its own clock, as curvature, reference speed, planned lateral and longitudinal acceleration, and grade at 10\,Hz.
A stack of dilated temporal blocks, conditioned on the start state, $\hat\xi_g$ and the prior's trajectory, predicts a correction to the prior's whole response series: speed, yaw rate, roll, tracking errors and the four wheel loads.
The correction has one output head per label fidelity, one trained on ROM-in-the-loop labels and one on the additional FLIP labels, and a final head predicts each peak, with an uncertainty, from the corrected series.
In total the model has about $1.3\times10^5$ learned weights.

\noindent\textbf{Batched evaluation.}
Only the state estimate depends on the history, and it is computed once per decision. The prior, a fixed-step recurrence of 100 steps compiled for deployment into one GPU kernel, and the feed-forward correction then score a decision's candidates, or those of many decisions, in one pass (\cref{sec:scale}).

We train it on two kinds of labels from the same vehicle: 10,320 runs simulated with the ROM in the loop and 480 FLIP runs, each giving 11 start points (decision times) along the run, plus a near-limit set of 1,800 ROM and 344 FLIP runs.
The prior is fitted first and frozen, and the correction is then trained by regression on the response series and the peaks.
For decisions in the standard tank we use an ensemble of five seeds with the ROM-fidelity head.
In the wide tank no nonlinear ROM could be identified, so the state estimate and the prior use the linear theory ROM of that tank. The prior is refitted on wide-tank ROM runs and refined on FLIP training runs, the correction is retrained on wide-tank data drawn from 17,400 ROM and about 1,050 FLIP runs, and decisions use a three-seed ensemble with the FLIP-fidelity head, all fixed before the wide-tank evaluation set existed.
We denote this model C$_c$.

\section{Experiments}
\label{sec:experiments}

\subsection{Experimental Setup}
\label{sec:setup}
\noindent\textbf{Evaluation sets.}
None of the roads or plans of the 120 evaluation groups appear in training or model selection, which we verify by hashing every standard-tank plan. The wide-tank roads are new.
In the standard tank we use 60 groups with 643 candidate plans on new roads of six kinds.
In the wide tank we use 60 groups on 12 new roads, designed to isolate the liquid state.
Each road is driven once with the liquid settled and once after each of four preceding maneuvers (\emph{pasts}) that leave the liquid moving (\cref{fig:example}, top). In the past groups the vehicle state at the decision point is copied exactly from the settled run, so for the reference the groups of a road differ only in their liquid state, and the models also see the run-up that produced it.
Each group has 16 candidates, four speed levels around the safety limit at four start delays, which gives 48 matched contrasts from 1,176 FLIP runs. An earlier set built the same way adds 16 contrasts but was also used to choose C$_c$'s prior.
Repeating fixed candidates in identical settings (64 repeat runs) changes peak LTR by a median 0.006--0.007 and at most 0.03 through run-to-run nondeterminism, and never changes a label.

\noindent\textbf{Baselines.}
Besides our model C and its ablations without the physics prior (L) and without the liquid-state input (N), we compare against planners that differ only in the predictor $f$ of \cref{eq:select}.
E restarts the multibody vehicle with the ROM liquid from the reference's start state and simulates every candidate, and R does the same with the liquid frozen as a rigid payload. Both fade the reference's measured slosh load into their own over the first second, so R still sees the moving liquid for about 1\,s.
B is our physics prior alone. In the wide tank, B$_c$ is the refitted prior of C$_c$, and E$_w$ runs E with that tank's linear-theory ROM.
Q is a nonlinear pendulum after Qi \etal~\cite{qi_anti-rollover_2025}, with its swing equation derived from the positions and energies in the authors' report and its parameters identified by differential evolution from five FLIP step responses per axis and tank.
K is a Kolaei-type linear potential-flow multimodal model~\cite{kolaei_coupled_2017,abramson_dynamic_1966} with three modes per axis, all constants from the rectangular tank's geometry, and damping from Kolaei \etal's empirical relation with its constant set to 1.
Q and K replace only the liquid inside the vehicle prior of B and start from the same liquid-state estimate as the other planners.
In the standard tank we also compare a quasi-static liquid, a linear equivalent mechanical model with our ROM's constants, and a Gaussian-process residual on the prior~\cite{kabzan_learning-based_2019}, with results in \cref{sec:supp-tau}.

\noindent\textbf{Operating points and statistics.}
At the raw operating point, $b=0$ in \cref{eq:select}.
At the calibrated point, $b$ is each planner's own mean bias in peak LTR, measured on separate reference runs that share no group with the evaluation, which removes any constant offset a planner could know in advance.
Two planners are compared on the groups where exactly one of them makes an avoidable unsafe choice, with a one-sided exact sign test, Holm-corrected within each family.
Prediction error is the mean absolute error (MAE) of the uncalibrated peak-LTR prediction, over near-limit plans ($\ell\in[0.4,0.8]$) unless stated, with 95\,\% intervals from a bootstrap over road groups.
Each test we call pre-registered was declared in a written protocol whose hash, fixing its scoring code, was recorded before the set's outcomes were computed. The wide-tank set was scored once, while the standard-tank set served several successive studies. A few earlier tests, which we call pre-declared, had written protocols without hashes. \Cref{sec:supp-prereg,sec:supp-deviations} list all protocols, deviations and post-hoc analyses.

\begin{table}[t]
\centering
\small
\setlength{\tabcolsep}{3.2pt}
\resizebox{\linewidth}{!}{%
\begin{tabular}{@{}l rrr rrr@{}}
\toprule
 & \multicolumn{3}{c}{Standard tank} & \multicolumn{3}{c}{Wide tank} \\
\cmidrule(lr){2-4}\cmidrule(l){5-7}
Planner & Unsafe$\downarrow$ & Best$\uparrow$ & MAE$\downarrow$ & Unsafe$\downarrow$ & Best$\uparrow$ & MAE$\downarrow$ \\
\midrule
C / C$_c$ (ours)        & \textbf{0} & 52 & .024 & \textbf{11} & 14 & .037 \\
E / E$_w$ (ROM in the loop) & 3 & 55 & .012 & 21 & 26 & .088 \\
R (rigid payload)         & 9 & 49 & .051 & 27 & 21 & .284 \\
\midrule
Q, after Qi \etal~\cite{qi_anti-rollover_2025} & 22 & 35 & .052 & 21 & 4 & .106 \\
K, Kolaei-type~\cite{kolaei_coupled_2017} & 14 & 43 & .050 & 24 & 11 & .108 \\
B / B$_c$ (prior alone) & 17 & 38 & .049 & 17 & 11 & .097 \\
\bottomrule
\end{tabular}}
\caption{Decisions at the calibrated operating point, $\tau=0.6$, 60 groups per tank. \emph{Unsafe}: avoidable unsafe choices. \emph{Best}: groups in which the planner picks the reference's best safe plan. \emph{MAE}: error of the uncalibrated peak-LTR prediction over all evaluation plans. In the wide tank, E$_w$ and C$_c$ use a linear-theory ROM that was not identified on that tank, and E$_w$ is a reported control, not a tested arm. Further standard-tank baselines are in \cref{tab:supp-extra}.}
\label{tab:main}
\end{table}

\begin{figure}[t]
  \centering
  \includegraphics[width=\linewidth]{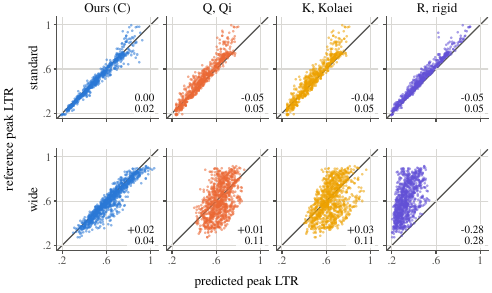}
  \caption{Predicted against reference peak LTR for every evaluation plan, before calibration (643 plans in the standard tank, 960 in the wide tank), with bias and MAE in each panel. Points above the diagonal are under-predictions, and dashed lines mark $\tau=0.6$. In the standard tank the published models and rigid payload under-predict by 0.04 to 0.05 on average but rank plans well. In the wide tank, the published models scatter widely, and rigid payload under-predicts every plan, by 0.28 on average.}
  \label{fig:scatter}
\end{figure}

\subsection{Scalability}
\label{sec:scale}
\begin{figure}[t]
  \centering
  \includegraphics[width=\linewidth]{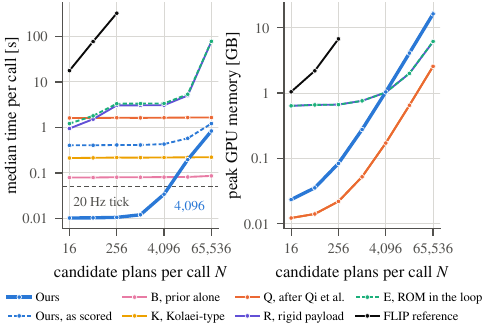}
  \caption{Scaling on one GPU (pre-registered). Left: median time per call against the number of candidate plans per call. Below the dashed line, a call fits one 20\,Hz tick. Right: peak GPU memory. Our deployed model fits the tick up to 4,096 plans, while its as-scored implementation, the physics-only planners and every simulator in the loop miss it at every batch size. FLIP stops at 256 plans (median call over 300\,s). In memory, B, K and Q coincide, as do R and E and the two implementations of ours.}
  \label{fig:scale}
\end{figure}
\Cref{fig:scale} reports a pre-registered benchmark that times every planner of \cref{tab:main} and the FLIP reference on batches of 16 to 65,536 candidate plans, from inputs on the GPU to safety flags on the host, together with 1,000 consecutive decisions at the size of a decision group and checks of numerical stability (\cref{sec:supp-cost}).
It ran on one NVIDIA RTX 6000 Ada, although the protocol named an H200.
Our model scores 4,096 plans within a 20\,Hz tick (99th-percentile time per call at most 50\,ms). E and R, which simulate the multibody vehicle for every candidate, and FLIP fit no batch size in a tick. Even at the smallest batch, 16 plans, a call takes 1.2, 0.94 and 18\,s.
At $N=65{,}536$ our model's throughput is 93$\times$ E's (95\,\% CI 93.3--93.5), and at $N=256$, the largest batch FLIP reached, 30,000$\times$ FLIP's.
Over 1,000 decisions its 99th-percentile latency is 11.2\,ms, with none above one tick.
Its predictions are also stable. The same plan's peak LTR changes by at most $4.3\times10^{-5}$ across batch sizes and positions in the batch, no safety flag changes, and no output is non-finite.
The physics-only planners, run without the compiled kernel, are cheaper per plan at large batches but miss the tick by a fixed cost per call (\cref{sec:supp-cost}).
The timed implementation compiles the prior into a GPU kernel. On this GPU its peak-LTR predictions differ from the stored scored ones by at most $5.2\times10^{-3}$, below the reference's median repeat noise (0.0062), with identical decisions on all 321 plans. This exceeds the protocol's $10^{-3}$ tolerance, so the agreement check fails. The implementation that produced the scored decisions takes 404\,ms per decision.

\subsection{Comparison with Physics-Based Planners}
\label{sec:vsphysics}
\Cref{tab:main} summarizes the decisions at the calibrated operating point, and \cref{fig:scatter} shows the predictions behind them.
Against the two published slosh models, all four pre-registered tests favor our model. It makes 0 avoidable unsafe choices against 22 for Q and 14 for K in the standard tank (Holm $p=9.5\times10^{-7}$ and $1.8\times10^{-4}$), and 11 against 21 and 24 in the wide tank ($p=0.015$ and $0.007$).
Like B, Q and K are physics-only planners without a learned correction.
The margins arise differently in the two tanks (\cref{fig:scatter}).
In the standard tank the held-out calibration shifts Q and K by less than 0.01, while on the evaluation plans they remain about 0.04 too low. Most of the margin is therefore a load bias on new roads that calibration from separate runs does not remove, and that our model avoids.
In the wide tank Q is nearly unbiased, but its error over all plans is about three times ours (MAE 0.106 against 0.037), so the margin there is accuracy. On the same groups, Q matches the simulation-based planner E$_w$ (21 each) and K is close (24).
Across thresholds our model keeps both decision metrics in a narrow band in both tanks, while the other fast models move with the threshold, trading unsafe choices for over-caution (\cref{sec:supp-tau}).
Our physics prior B alone makes 17 under the same calibration. The difference is prediction accuracy (\cref{tab:main}), which also makes our model less sensitive to the threshold (\cref{sec:supp-tau}).

Against E, which simulates the reduced-order liquid with the multibody vehicle for every candidate, our model is non-inferior. Both make 3 avoidable unsafe choices at the raw threshold in the standard tank, and in a pre-registered wide-tank test on 20 groups our model met the non-inferiority bound (superiority $p=0.06$).
It reaches these decisions at a fraction of E's cost (\cref{sec:scale}).

\begin{figure}[t]
  \centering
  \includegraphics[width=\linewidth]{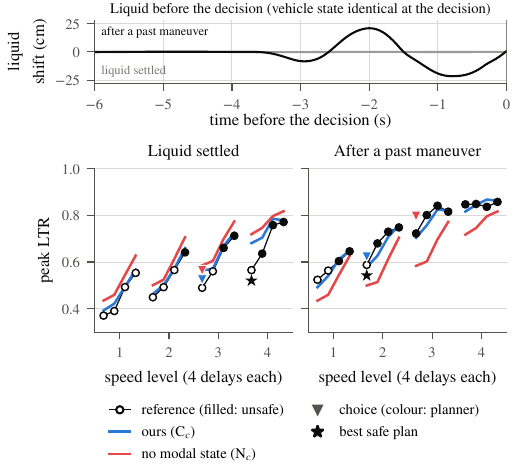}
  \caption{One wide-tank contrast: the same road and identical vehicle state, with the liquid settled (left) or carrying a past maneuver (right). Top: the FLIP liquid's lateral shift before the decision in the two groups. Bottom: the 16 candidates' reference peak LTR and the calibrated predictions, and triangles mark each planner's choice. Without the liquid state, N$_c$ predicts the same loads in both cases and keeps its choice, which becomes unsafe, while C$_c$ raises its predictions and switches to the best safe plan. It is one of 7 contrasts in which the reference's best plan changes and only N$_c$'s choice becomes avoidably unsafe. In the settled case both planners are more cautious than necessary.}
  \label{fig:example}
\end{figure}

\subsection{Effect of Liquid Motion and State}
\label{sec:state}
A planner that ignores the liquid's motion makes many avoidable unsafe choices.
In the standard tank at the raw threshold, the rigid-payload planner R makes 23 avoidable unsafe choices in 60 groups and E makes 3, with all 20 discordant groups against R ($p=9.5\times10^{-7}$), and the direction holds for every $\tau$ between 0.5 and 0.7. Calibration removes most but not all of this gap. At the calibrated operating point, R still makes 9 against E's 3 (7:1, $p=0.035$, a pre-declared secondary test). At the raw threshold, a quasi-static liquid inside the reduced vehicle prior, which keeps the liquid's shift but not its motion, makes 28.
R ranks plans well (AUROC 0.99 or higher over all plans) but under-predicts near-limit peak LTR by a nearly uniform 0.055, so a fixed threshold calls too many plans safe.
In the wide tank its near-limit offset grows to 0.27, yet its excess over E$_w$ (own bias $-0.05$) was not significant in a post-hoc test (10:4, $p=0.09$ raw and 12:6, $p=0.12$ calibrated).

\begin{table}[t]
\centering
\small
\setlength{\tabcolsep}{4pt}
\resizebox{\linewidth}{!}{%
\begin{tabular}{@{}l c c c c@{}}
\toprule
Planner & Near-limit MAE & Follows past & Unsafe & Best \\
\midrule
C$_c$ (ours)              & 0.036 {\scriptsize[.032, .041]} & 28 & 11 & 14 \\
N$_c$ (no modal state)   & 0.079 {\scriptsize[.070, .089]} & 6  & 15 & 6 \\
L (no physics prior)      & 0.037 {\scriptsize[.032, .043]} & 24 & 15 & 36 \\
Reference                 & --                              & 34 & 0  & -- \\
\bottomrule
\end{tabular}}
\caption{The liquid state in the wide tank (60 groups, 48 contrasts). Near-limit MAE is of uncalibrated predictions, with 95\,\% bootstrap intervals. Decisions are calibrated, at $\tau=0.6$. \emph{Follows past}: contrasts in which the chosen plan differs between the settled and the past group, although the vehicle states are identical. N$_c$ has no modal state or latent but still reads recent load transfer from the observed history.}
\label{tab:state}
\end{table}

We next test whether the liquid's \emph{current state} changes decisions.
In the standard tank, no pre-declared test found a decision changed by the liquid state.
In the wide tank, with the vehicle state held identical, the carried-over liquid state alone changed the reference's best plan in 13 of 16 contrasts, mostly through the chosen speed level.
That test was rated noise-limited by its own pre-declared rule, because the median repeat noise (0.0062) exceeded its 0.005 bar, so we treat the effect as directional. A design check on the 60-group set agrees (\cref{sec:supp-sets}).
\Cref{tab:state} shows that our model reads this state, and \cref{fig:example} gives an example.
Removing the liquid-state input doubles near-limit error, from 0.036 to 0.079 with non-overlapping intervals, a gap five times the reference's repeat noise. With the input, the model's choice follows the past in 28 of 48 contrasts, against 6 without it and 34 for the reference.
Its decision gain was in the same direction but not significant. In the pre-registered comparison, 7 groups had an avoidable unsafe choice only without the state and 3 only with it ($p=0.17$), a test with power 0.23 at the observed effect. Confirming an effect of this size would need about 250 groups.
One reason is that C$_c$ is cautious and picks the best safe plan in only 14 of 60 groups (\cref{tab:state}).
Ablations of the physics prior, the output heads and the recurrent latent are in \cref{sec:prior}.

\section{Discussion and Conclusion}
\label{sec:conclusion}
Payload-carrying robots plan by predicting many candidate motions at once. In PLiMB, predicting them with the liquid as a rigid payload leads to many avoidable unsafe choices.
Our physics-anchored world model scores 4,096 candidate plans per control tick on one GPU and makes fewer avoidable unsafe choices than rigid-payload and published low-order slosh planners in both tanks, none in the standard tank. When the sloshing mode lies in the maneuver band, it reads the liquid state left by earlier driving from the vehicle's history, which halves its near-limit error.
The same structure, a batched physical prior, a state estimated from the carrier's own motion and a learned correction, may extend to other hidden, history-dependent payloads (granular loads, hanging masses) and to other carriers of liquid, which we have not tested.

The decision gains from the liquid state and from the physics prior pointed the right way but were not significant. Effects of the observed size would need roughly 250 and 300 groups.
We study one vehicle, two tank widths at one fill and one family of maneuvers with single plan selections. The hidden-state groups sit at the safety limit by design, and the published models were adapted to our vehicle prior. Our reference agrees with an independent CFD solver but still lacks a wide-tank grid study and experimental validation of the coupled vehicle.

{
    \small
    \bibliographystyle{ieeenat_fullname}
    \bibliography{references,addition}
}

\clearpage
\appendix

\section{Pre-registration Record}
\label{sec:supp-prereg}
\Cref{tab:supp-prereg} lists every protocol behind a result in the paper.
For the studies marked pre-registered, the protocol and any addendum were hashed (SHA-256) before the outcomes of the evaluation set were computed, which fixed the sets, operating points, statistics, readings and scoring code in advance.
The earliest studies had written protocols without hashes, and we call them pre-declared.
Deviations are listed in \cref{sec:supp-deviations}.

\begin{table*}[t]
\centering
\small
\resizebox{\linewidth}{!}{%
\begin{tabular}{@{}l l l l@{}}
\toprule
Study (where used) & Primary question & Protocol / addendum hash & Status \\
\midrule
Learned model vs.\ uncalibrated prior (\cref{sec:vsphysics}) & C vs.\ B, raw $\tau$, 30 groups & \texttt{91446bb6} & pre-registered \\
Shift and decision metrics, data efficiency (\cref{sec:prior}) & shift, data efficiency, decisions & \texttt{49d9d3b4}, refit \texttt{f83dd7a4} & pre-registered \\
Rigid payload (\cref{sec:state}) & R vs.\ E, 60 groups & \texttt{52b6af6e} & pre-registered \\
Controlled contrasts (\cref{sec:state}) & does the liquid state change the best plan? & \texttt{aed8fb1d} / \texttt{8fca0983} & pre-registered \\
Wide-tank step 3 (\cref{sec:vsphysics}) & C vs.\ E$_w$, 20 groups & \texttt{3d290b6c} / \texttt{52a3bce1} & pre-registered \\
Wide-tank confirmation (\cref{sec:state,sec:prior}) & C$_c$ vs.\ N$_c$, C$_c$ vs.\ L & \texttt{da14a2fa} / \texttt{3c7b7b2c} & pre-registered \\
Published baselines (\cref{sec:vsphysics}) & C vs.\ Q, K, both tanks & \texttt{b68893c9} / \texttt{b4c3940e} & pre-registered \\
Cost per decision, superseded by P14 & C vs.\ E, standard tank & \texttt{d302cdd9} / \texttt{abe5a964} & pre-registered \\
Scalability, P14 (\cref{sec:scale}) & all planners and FLIP, one GPU & \texttt{50ca3235} / \texttt{09c61b34}, \texttt{c7ea0e5a} & pre-registered, GPU deviation \\
Rigid payload, calibrated (\cref{sec:state}) & R vs.\ E, calibrated (secondary) & \texttt{e95b4a08} & pre-registered \\
Simulator validation (\cref{sec:supp-validation}) & FLIP vs.\ CFD and experiments & hashed protocol and addenda & pre-registered \\
Calibrated prior, FLIP head, state in standard tank & C vs.\ calibrated B, heads, history & -- & pre-declared \\
\bottomrule
\end{tabular}}
\caption{Protocols behind the results. Hashes are the first eight hex digits of the SHA-256 of the protocol file, recorded before scoring. The full files are part of the released material.}
\label{tab:supp-prereg}
\end{table*}

\section{Evaluation Sets and Operating Points}
\label{sec:supp-sets}
\noindent\textbf{Standard tank.}
The 60 groups come from three batches of fresh roads (12, 18 and 30 groups), each road driven by a run-up that leaves the liquid in motion before the decision point.
Candidates come from families that differ in speed level near the limit, braking point (late braking, braking in the curve) and path. Candidates that violate actuator, steering-rate or grip limits are dropped.
A hash of every plan's curvature and reference speed is checked against all training and model-selection rows, and no plan or road overlaps.

\noindent\textbf{Wide tank.}
Each of the 12 roads has one settled group and four past groups.
The pasts are lateral cycles of random amplitude, count, duration and sign, drawn with a fixed seed and kept away from any past used in earlier studies.
The four speed levels are set per road from a reduced-order sweep so that, under the settled liquid, about 0, 25, 50 and 75\,\% of the candidates are unsafe. In the reference, the realized shares were 0, 25, 50 and 79\,\%.
In each past group the vehicle state at the decision point is copied from the settled group's run, and the liquid and run-up record from a donor run that drove the past, so for the reference the two groups of a contrast differ only in the liquid, while the models observe the run-up record that produced it.
Repeating two fixed candidates per road in identical settings changes peak LTR by a median 0.0072 (maximum 0.030), with no label change.

\noindent\textbf{Calibration.}
The calibrated operating point subtracts each planner's mean signed error in peak LTR on held-out reference rows.
In the standard tank these are the development and near-limit selection rows of training, which share no group with the evaluation. In the wide tank, they are the bias rows declared in the confirmation protocol.
\Cref{tab:supp-shift} lists the shifts.
An earlier pre-declared test of the prior B used a different calibration, which instead lowers each planner's threshold to the value its own calibration rows support (\cref{sec:vsphysics}).

\begin{table}[t]
\centering
\small
\begin{tabular}{@{}l r l r@{}}
\toprule
\multicolumn{2}{c}{Standard tank} & \multicolumn{2}{c}{Wide tank} \\
\cmidrule(r){1-2}\cmidrule(l){3-4}
C & $-0.008$ & C$_c$ & $+0.007$ \\
E & $-0.005$ & N$_c$ & $-0.029$ \\
R & $-0.037$ & & \\
Q & $-0.007$ & L & $+0.022$ \\
K & $-0.006$ & E$_w$ & $-0.046$ \\
B & $-0.006$ & R & $-0.278$ \\
Quasi-static & $-0.016$ & Q & $-0.010$ \\
Linear equiv.\ mech. & $-0.007$ & K & $-0.001$ \\
GP residual & $-0.003$ & B$_c$ & $-0.043$ \\
\bottomrule
\end{tabular}
\caption{Calibration shift (mean predicted minus reference peak LTR on held-out rows) of each planner.}
\label{tab:supp-shift}
\end{table}

\noindent\textbf{Run inventory.}
\Cref{tab:supp-runs} lists every coupled vehicle run behind the paper, by tank and set.
Each run drives one road and plan. The evaluation sets also count the run-up and donor runs from which their groups are cloned, and the repeat runs.
Not counted are the simulator-validation study (\cref{sec:supp-validation}), the timing runs (\cref{sec:supp-cost}), and the candidates that the planners E and R simulate at decision time.

\begin{table}[t]
\centering
\small
\setlength{\tabcolsep}{3pt}
\resizebox{\linewidth}{!}{%
\begin{tabular}{@{}l l r r@{}}
\toprule
Set & Use & FLIP & ROM \\
\midrule
\multicolumn{4}{@{}l}{\emph{Standard tank}} \\
Training roads and speeds & training & 480 & 10,320 \\
Near-limit training & training & 344 & 1,800 \\
Development, near-limit selection & selection and calibration rows & 280 & 840 \\
Near-limit calibration & thresholds of earlier studies & 96 & 600 \\
Held-out roads and speeds & earlier studies & 854 & 960 \\
Second seed of the training suite & repeatability & 202 & -- \\
Evaluation, 60 groups (643 candidates) & \cref{tab:main} & 703 & -- \\
Step responses for Q & identification & 10 & -- \\
\emph{Subtotal} & & \emph{2,969} & \emph{14,520} \\
\midrule
\multicolumn{4}{@{}l}{\emph{Wide tank}} \\
Screening study & 154 reused in training & 159 & -- \\
Training and selection & training & 896 & 17,400 \\
Earlier set, 20 groups & non-inferiority test & 424 & -- \\
Evaluation, 60 groups (960 candidates) & \cref{tab:main,tab:state} & 1,176 & -- \\
Step responses for Q & identification & 10 & -- \\
\emph{Subtotal} & & \emph{2,665} & \emph{17,400} \\
\midrule
\emph{Total} & & \emph{5,634} & \emph{31,920} \\
\bottomrule
\end{tabular}}
\caption{Coupled vehicle runs by tank and set, with the liquid simulated by FLIP or by the reduced-order model (ROM). Wide-tank evaluation sets include donor runs (64 and 192) and repeat runs (40 and 24).}
\label{tab:supp-runs}
\end{table}

\noindent\textbf{Hidden-state design check (\cref{sec:state}).} In the 60-group wide-tank set, whose repeat noise also exceeds the pre-declared bar and which is not an independent test, the best plan differs between the settled and the past group in 34 of 48 contrasts, against about 12\,\% in a post-hoc noise-only placebo.

\section{Results at Every Threshold}
\label{sec:supp-tau}
\Cref{tab:supp-tau} gives the avoidable unsafe choices of every planner for $\tau$ from 0.50 to 0.70, calibrated and raw.
In the standard tank our model's margin over the published models and the physics baselines holds at every threshold.
In the wide tank it is largest at high thresholds and narrows at $\tau=0.50$, where the prior alone (B$_c$) makes fewer avoidable unsafe choices.
In the liquid-state comparison, the model with the state (C$_c$) makes fewer avoidable unsafe choices than N$_c$ at every threshold, although the pre-registered sign test, at $\tau=0.6$ only, was not significant, and the secondary looks are significant at 0.65 and 0.70.

\begin{table}[t]
\centering
\small
\setlength{\tabcolsep}{3pt}
\resizebox{\linewidth}{!}{%
\begin{tabular}{@{}l ccccc ccccc@{}}
\toprule
 & \multicolumn{5}{c}{Calibrated, $\tau=$} & \multicolumn{5}{c}{Raw, $\tau=$} \\
\cmidrule(lr){2-6}\cmidrule(l){7-11}
Planner & .50 & .55 & .60 & .65 & .70 & .50 & .55 & .60 & .65 & .70 \\
\midrule
\multicolumn{11}{@{}l}{\emph{Standard tank}} \\
C (ours) & 5 & 3 & 0 & 6 & 5 & 9 & 5 & 3 & 7 & 6 \\
E (ROM in the loop) & 5 & 5 & 3 & 4 & 2 & 10 & 8 & 3 & 6 & 5 \\
R (rigid payload) & 13 & 16 & 9 & 12 & 9 & 25 & 23 & 23 & 19 & 13 \\
Q & 23 & 22 & 22 & 22 & 23 & 26 & 22 & 24 & 25 & 23 \\
K & 22 & 18 & 14 & 22 & 23 & 22 & 19 & 17 & 22 & 23 \\
B (prior alone) & 22 & 17 & 17 & 21 & 23 & 24 & 19 & 20 & 23 & 23 \\
Quasi-static & 23 & 20 & 18 & 25 & 23 & 26 & 24 & 28 & 27 & 23 \\
Linear equiv.\ mech. & 22 & 18 & 13 & 21 & 23 & 23 & 19 & 19 & 22 & 23 \\
Linear equiv., step-identified & 24 & 19 & 16 & 22 & 23 & 25 & 20 & 22 & 22 & 23 \\
GP residual & 17 & 14 & 13 & 9 & 9 & 17 & 16 & 13 & 11 & 9 \\
\midrule
\multicolumn{11}{@{}l}{\emph{Wide tank}} \\
C$_c$ (ours) & 9 & 9 & 11 & 7 & 5 & 8 & 8 & 8 & 6 & 5 \\
N$_c$ (no modal state) & 10 & 12 & 15 & 15 & 18 & 15 & 17 & 18 & 19 & 21 \\
L (no physics prior) & 22 & 26 & 15 & 11 & 8 & 18 & 20 & 11 & 8 & 7 \\
E$_w$ (ROM in the loop) & 22 & 30 & 21 & 17 & 11 & 33 & 39 & 26 & 19 & 11 \\
R (rigid payload) & 12 & 25 & 27 & 27 & 21 & 47 & 45 & 32 & 27 & 21 \\
Q & 11 & 16 & 21 & 28 & 25 & 13 & 19 & 24 & 31 & 25 \\
K & 14 & 15 & 24 & 26 & 23 & 14 & 15 & 24 & 26 & 23 \\
B$_c$ (prior alone) & 4 & 10 & 17 & 26 & 21 & 9 & 20 & 28 & 25 & 21 \\
\bottomrule
\end{tabular}}
\caption{Avoidable unsafe choices (of 60 groups) at every threshold, at the calibrated and raw operating points. Only $\tau=0.6$ at the calibrated point was a primary test, and all other cells are pre-declared secondary analyses.}
\label{tab:supp-tau}
\end{table}

\Cref{tab:supp-extra} gives the four further baselines of the standard tank in the format of \cref{tab:main}. All four make more avoidable unsafe choices than our model at every threshold (\cref{tab:supp-tau}).

\begin{table}[t]
\centering
\small
\begin{tabular}{@{}l rrr@{}}
\toprule
Planner (standard tank) & Unsafe$\downarrow$ & Best$\uparrow$ & MAE$\downarrow$ \\
\midrule
Quasi-static liquid & 18 & 39 & .061 \\
Linear equiv.\ (ROM constants) & 13 & 43 & .050 \\
Linear equiv., step-identified & 16 & 40 & .050 \\
GP residual~\cite{kabzan_learning-based_2019} & 13 & 37 & .043 \\
\bottomrule
\end{tabular}
\caption{Further standard-tank baselines at the calibrated operating point, $\tau=0.6$, 60 groups (our model: 0, 52, .024).}
\label{tab:supp-extra}
\end{table}

\noindent\textbf{Trade-off across thresholds.}
\begin{figure}[t]
  \centering
  \includegraphics[width=\linewidth]{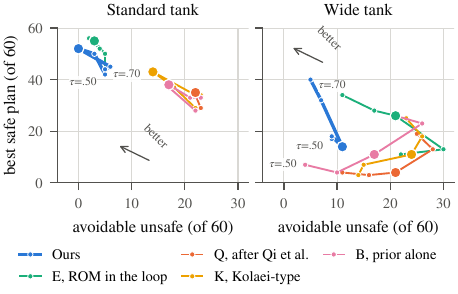}
  \caption{Both decision metrics across thresholds at the calibrated operating point (60 groups per tank). Each line joins $\tau=0.5,0.55,\dots,0.7$. The large marker is the pre-registered $\tau=0.6$, the other thresholds are pre-declared secondary analyses. A planner can avoid unsafe choices by caution, at the cost of rarely picking the best safe plan, so a good model sits in the upper left.}
  \label{fig:tau}
\end{figure}

Across thresholds (\cref{fig:tau}), our model stays in the upper left in both tanks, with 5 to 11 avoidable unsafe choices and 14 to 40 best safe plans in the wide tank. The other fast models move with the threshold instead. In the wide tank at $\tau=0.5$, the prior alone makes fewer avoidable unsafe choices than our model (4 against 9) and Q is close (11, with 7:5 discordant groups), but they pick the best safe plan in only 7 and 4 groups against our 17, because at a low threshold they fall back on the slowest plans. Post hoc, Q's and K's predictions within a group are compressed (spread 0.69 and 0.73 of the reference's, Spearman 0.71 and 0.68 against our 0.94), so the threshold, more than the plan, sets their choice.
The prior B alone shows how much of such a margin is a constant offset.
Under the held-out shift of \cref{tab:supp-shift}, B makes 17 avoidable unsafe choices against our model's 0, and at the raw threshold on 30 fresh groups it made 12 against our model's 2 ($p=0.003$, no loss of progress).
Most of the raw gap is B's under-prediction. In an earlier pre-declared test on 18 groups, in which B's threshold was calibrated on the same groups, the two planners differed in a single group.
Our model's gain over B is thus mainly in accuracy, with near-limit error falling from 0.053 to 0.023 in that test, which makes it less sensitive to the threshold.

\section{Published Baselines as Implemented}
\label{sec:supp-baselines}
Both published models replace only the liquid inside the vehicle prior of B and start from the liquid-state estimate of \cref{sec:model}, so that every planner shares the same vehicle model and information.

\noindent\textbf{Q, after Qi \etal~\cite{qi_anti-rollover_2025}.}
A fixed mass $m_0=M-m_p$ at height $h_0$ and a swing mass $m_p$ on a pendulum of length $l_p$ with viscous damping $c_d$.
The swing equation is derived from the positions and energies in the authors' report, with the lateral acceleration and the roll angle, rate and acceleration of the prior's vehicle as inputs. The loads are the exact Newton--Euler loads of the two point masses.
The vehicle receives Q's total liquid load minus the frozen load of the settled liquid, which the chassis already carries.
The five parameters per axis are identified by differential evolution (population 30, at most 300 generations) on five FLIP step responses per axis and tank, of 1--5\,m/s$^2$ held for 10\,s, minimizing the variance-normalized error of the lateral force, vertical force and roll moment, in place of the CFD used by the authors.
\Cref{tab:supp-q} gives the identified parameters.
A second pendulum on the longitudinal axis is identified the same way. Because our prior reads only the liquid's roll moment, it does not change any prediction.
The initial pendulum state is mapped from the estimated displacement and velocity of the liquid's center of mass.
Fitting with the authors' mean-normalized objective instead did not improve Q (21 and 25 avoidable unsafe choices in the standard and wide tanks, against 22 and 21, post hoc).

\begin{table}[t]
\centering
\small
\setlength{\tabcolsep}{3pt}
\resizebox{\linewidth}{!}{%
\begin{tabular}{@{}l c c c c c c@{}}
\toprule
Tank, axis & $m_p/M$ & $h_0$ [m] & $h_p$ [m] & $l_p$ [m] & $c_d$ [s$^{-1}$] & $\zeta$ \\
\midrule
Standard, lateral      & 0.599 & 0.293 & 0.103 & 0.418 & 0.471 & 0.049 \\
Standard, longitudinal & 0.610 & 0.333 & 0.115 & 0.541 & 0.400 & 0.047 \\
Wide, lateral          & 0.685 & 0.392 & 0.092 & 0.769 & 0.427 & 0.060 \\
Wide, longitudinal     & 0.636 & 0.575 & 0.081 & 0.978 & 0.272 & 0.043 \\
\bottomrule
\end{tabular}}
\caption{Identified parameters of Q. $M$ is the liquid mass (480\,kg standard, 1,050\,kg wide).}
\label{tab:supp-q}
\end{table}

\noindent\textbf{K, a Kolaei-type model~\cite{kolaei_coupled_2017}.}
The linear potential-flow equivalent of a rectangular tank~\cite{abramson_dynamic_1966,dodge_new_nodate}, with the first three antisymmetric modes per axis. Frequencies, modal masses, heights and the fixed mass follow from the tank geometry and fill, with nothing fitted.
Damping uses the empirical relation $\xi=C_1\sqrt{\nu/(g^{1/2}(2a)^{3/2})}$ with $C_1=1$ for a rectangular tank, giving $\xi\approx4$--$6\times10^{-4}$ for water.
Higher modes start at rest, and mode 1 from the estimated liquid state.
In the standard tank the first-mode frequencies are 5.12 and 4.48\,rad/s (lateral, longitudinal) and in the wide tank 3.75 and 3.29\,rad/s.

\section{Scalability Benchmark}
\label{sec:supp-cost}
The scaling numbers of \cref{sec:scale} come from P14, a pre-registered benchmark (protocol \texttt{50ca3235}, addendum \texttt{09c61b34}) run by a coworker from a standalone bundle on one NVIDIA RTX 6000 Ada (48\,GB). The protocol named an H200. An addendum written after the results were seen (\texttt{c7ea0e5a}) records this deviation, makes this, the only completed run, the reported one, and records the agreement check below as not passed. No other process was on the GPU during any timed call.
It times C (deployed and as-scored paths), B, Q, K, R, E and the FLIP reference on the 321 plans of 30 of the 60 standard-tank evaluation groups, tiled to batches of $N=16$ to 65,536, with three untimed warm-up calls and 50 timed calls per arm and $N$ (10 for R and E, and 5 for FLIP or 3 once a call exceeds 60\,s). An arm stops at the first $N$ whose median call exceeds 300\,s or that would leave less than 8\,GB of GPU memory free.
A call scores a whole batch, between two device synchronizations, from inputs on the GPU to safety flags on the host. For E and R, it includes restoring every vehicle's start state and the 5\,s multibody rollout, and FLIP is timed from a settled start because its particle states are not stored.
Latency stability is measured over 1,000 consecutive decisions at the size of a decision group (100 for Q, R and E, 20 for FLIP).
Numerical stability is checked by the count of non-finite outputs, by the change of each plan's prediction across batch sizes and positions in the batch, by repeatability across calls, and by agreement with the scored peaks.
\Cref{tab:p14} gives the results per planner.
The physics-only planners B, K and Q run the prior in PyTorch without the compiled kernel. At $N=65{,}536$, B and K take 0.10$\times$ and 0.27$\times$ our model's time per plan, but each call has a fixed cost of 79 and 212\,ms, and Q's sub-stepped pendulum takes 1.6\,s per call, so none fits a tick.
Agreement with the scored peaks at the size of a decision group: B, K and Q within $5.5\times10^{-6}$, R and E within $1.9\times10^{-4}$ with all flags identical (passed), and our model within $5.2\times10^{-3}$ (deployed) and $1.4\times10^{-3}$ (as scored), flags identical on all 321 plans, above the $10^{-3}$ tolerance (not passed). The scored peaks were computed on the evaluation GPU (RTX PRO 6000 Blackwell), where the same check gave $3.7\times10^{-3}$ and $1.1\times10^{-6}$.
FLIP's repeat difference, $5.2\times10^{-3}$, is of the size of its own repeat noise (median 0.0062).
\begin{table}[t]
\centering
\caption{P14 per planner, one RTX 6000 Ada. Plans per tick: largest tested $N$ with 99th-percentile time per call at most 50\,ms. Throughput at $N=65{,}536$ and peak memory at $N=4{,}096$ ($^\dagger$FLIP: at $N=256$, its largest). Latency: over 1,000 consecutive decisions (100 for Q, R and E, and 20 for FLIP), each at the size of a decision group.}
\label{tab:p14}
\resizebox{\linewidth}{!}{%
\begin{tabular}{lrrrrrl}
\toprule
Planner & plans/tick & plans/s & GB & p99 [ms] & $>$50\,ms & stability \\
\midrule
Ours (deployed) & 4{,}096 & 78{,}641 & 1.04 & 11.2 & 0/1000 & stable \\
Ours (as scored) & 0 & 54{,}007 & 1.06 & 416.5 & 1000/1000 & stable \\
B, prior alone & 0 & 762{,}861 & 0.17 & 81.1 & 1000/1000 & stable \\
K, Kolaei-type & 0 & 295{,}955 & 0.17 & 216.5 & 1000/1000 & stable \\
Q, after Qi et al. & 0 & 39{,}808 & 0.17 & 1{,}660.1 & 100/100 & stable \\
R, rigid payload & 0 & 846 & 1.02 & 937.4 & 100/100 & bitwise-stable \\
E, ROM in the loop & 0 & 842 & 1.02 & 1{,}193.0 & 100/100 & bitwise-stable \\
FLIP reference & 0 & 0.79$^\dagger$ & 6.76$^\dagger$ & 12{,}156.7 & 20/20 & noisy \\
\bottomrule
\end{tabular}}
\end{table}
An earlier pre-registered measurement on a shared RTX PRO 6000 (P12, \texttt{d302cdd9}) timed one decision group per call for C and E only, and is superseded by P14. Its 15\,ms and 126$\times$ were measured on another GPU and are not combined with P14's numbers.

\section{Positioning of PLiMB}
\label{sec:supp-position}
\Cref{tab:supp-position} compares PLiMB (Parallel Liquid--MultiBody), our GPU-parallel slosh simulator, with the simulators and coupled vehicle models closest to it, by what each cited work shows.
High-fidelity multibody and fluid--structure engines model the liquid with the vehicle but are not shown running many independent worlds per GPU. Batched GPU engines run many worlds but not with vehicle dynamics. The coupled vehicle models of slosh research use a low-order liquid in a single simulation.
PLiMB combines two-way FLIP coupling, multibody vehicles and many independent worlds on one GPU, each on its own road and plan. This combination makes a high-fidelity outcome for every candidate plan affordable.
\begin{table*}[t]
\centering
\small
\begin{tabular}{@{}l l c c c@{}}
\toprule
Work & Liquid model & Liquid--body coupling & Vehicle dynamics & Many worlds per GPU \\
\midrule
Chrono~\cite{tasora_chrono_2016} & SPH, potential flow & two-way & yes & -- \\
Rewarped~\cite{xing_stabilizing_2025} & MPM & one-way & -- & yes \\
Alloy~\cite{gupta_alloy_2026}$^\ast$ & SPH & two-way & -- & yes \\
Kolaei \etal~\cite{kolaei_coupled_2017} & linear multimodal & two-way & yes & -- \\
Qi \etal~\cite{qi_anti-rollover_2025} & equivalent pendulum & two-way & yes & -- \\
\midrule
PLiMB (ours) & FLIP & two-way & yes & yes \\
\bottomrule
\end{tabular}
\caption{PLiMB against the closest simulators and coupled vehicle models. ``yes'': shown in the cited work. ``--'': not shown there. $^\ast$Concurrent work. Coupled vehicle models are listed as published. Chrono's FSI module also couples to other fluid solvers.}
\label{tab:supp-position}
\end{table*}

\section{Simulator Validation}
\label{sec:supp-validation}
The validation study was pre-registered separately and used the solver code unchanged.
\emph{V2, against CFD:} the wide tank's lateral cross-section was driven, one-way, by the tank motions logged for six vehicles in the reference, and FLIP at the evaluation settings was compared with 2D interFoam on slosh force and roll moment: median NRMSE 0.053 and 0.047 (0.047 and 0.045 against a 1\,cm CFD mesh), median phase $-4.0^\circ$ and $-2.2^\circ$, and a static LTR shift at the vehicles' peaks of at most 0.035, against a median decision margin of 0.35. The first-mode frequency agrees with CFD and linear theory (0.599, 0.600 and 0.597\,Hz). The pre-registered run had used a neighboring FLIP configuration (ghost-fluid mode 1, more solver iterations) and gave the same class (0.055 and 0.046). We repeated it at the evaluation settings under a hashed addendum.
\emph{V1, SPHERIC Test 10~\cite{delorme_set_2009,souto-iglesias_canonical_part0,souto-iglesias_repeatability_2011,botia-vera_three_2010}:} lateral impact pressure in a rectangular tank under prescribed roll. FLIP's impact timing matches the measurement (mean phase $-5^\circ$ for water), and its per-cycle wall-pressure impulse is $1.83\times$ the measured one. Given the same motion, interFoam gives $2.0\times$, and the two solvers agree within 8\,\%, so the gap is shared by both solvers rather than specific to FLIP.
\emph{V3, SPHERIC Test 9~\cite{bulian_sph_2010,botia-vera_three_2010,bouscasse_mechanical_2014a,bouscasse_mechanical_2014b}:} a freely rolling tuned liquid damper, simulated two-way coupled. The peak roll with water is 37\,\% above the measurement with a $28^\circ$ lag, consistently across grids and seeds.

\section{Ablation Study}
\label{sec:prior}
In the standard tank, trained on 10\,\% of the training roads (321 fresh plans, three seeds), the model without the prior has a peak-LTR MAE higher by 0.016 [0.010, 0.024], and by 0.019 [0.012, 0.026] when the prior is refitted on the same subset (one subset draw).
On decisions, in the wide tank, the prior was non-inferior to no prior (11 against 15 avoidable unsafe choices, upper bound 0.017 against a margin of 0.15 per group) but not superior ($p=0.19$), and a maximally cautious planner would also pass the non-inferiority test. The model with the prior gives up 1.45\,m of progress against the one without. An earlier pre-registered superiority test on 20 wide-tank groups also failed, with the no-prior model ahead at the raw threshold (1 against 5).
The FLIP-fidelity output head likewise changed no decision in two pre-declared tests, and the recurrent latent added little over the ROM's modal state.

\section{Other Variants Explored}
\label{sec:supp-negative}
The following variants did not improve on our model.
A particle belief over liquid states screened plans no better than the model's own prediction (AUROC 0.777 against 0.779).
Training on interventional pairs that differ only in the liquid reduced the error of the predicted liquid effect by 9.9\,\%, short of the 15\,\% we required to adopt it.
Re-fitting the prior's constants and learning inside a closed planning loop did not improve on the open-loop model.
The FLIP-fidelity head and the recurrent latent are discussed in \cref{sec:prior}.

\section{Power}
\label{sec:supp-power}
\Cref{tab:supp-power} lists the exact power of the main sign tests and the number of groups an 80\,\% powered test would need, as a guide for follow-up studies.

\begin{table}[t]
\centering
\small
\setlength{\tabcolsep}{3pt}
\begin{tabular}{@{}l c c c@{}}
\toprule
Comparison & Split & Power & Groups for 80\,\% \\
\midrule
R vs.\ E, standard & 20:0 & $\approx1.0$ & 24 \\
C vs.\ uncalibrated B & 11:1 & 0.91 & 24 \\
C vs.\ Q / K, standard & 22:0 / 14:0 & 1.00 / 0.99 & -- \\
C$_c$ vs.\ Q / K, wide & 14:4 / 17:4 & 0.48 / 0.70 & -- \\
C$_c$ vs.\ N$_c$, wide & 7:3 & 0.23 & $\approx250$ \\
C$_c$ vs.\ L (superiority) & 4:8 & 0.22 & $\approx300$ \\
C vs.\ calibrated B & 1:0 & $\approx0$ & $\approx150$ \\
\bottomrule
\end{tabular}
\caption{Power of the main one-sided sign tests at the observed discordant split (60 groups unless stated, 30 for C vs.\ B and 18 for calibrated B). Power for Q and K is at the Holm-adjusted level.}
\label{tab:supp-power}
\end{table}

\section{Deviations}
\label{sec:supp-deviations}
The protocols were followed except as listed here. All were recorded in addenda before scoring unless marked post hoc.
\begin{itemize}
\item Published baselines: Q was identified on the 2.5\,cm grid of the data runs rather than the 4\,cm stated in the protocol. Its fitting objective normalizes by variance where the authors' normalizes by mean (checked post hoc, \cref{sec:supp-baselines}). Q and K are sub-stepped at 1/240\,s.
\item Cost per decision (P12): before the run, an addendum made the deployed path primary and set the correctness tolerance to $10^{-3}$ (registered $10^{-5}$). After the run, we accepted the deployed path's $3.7\times10^{-3}$ difference, which left every decision unchanged.
\item Wide-tank existing planners' lane errors were recomputed for the published-baseline study (reproduced to $5\times10^{-8}$), and its bootstrap was clustered by road rather than by road and past, which widens the intervals.
\item Scalability (P14, post hoc): the run used an RTX 6000 Ada instead of the H200 named in the protocol, and our model's agreement with the scored peaks ($5.2\times10^{-3}$ deployed, $1.4\times10^{-3}$ as scored) exceeded the $10^{-3}$ tolerance, so the check is reported as not passed, with identical decisions (\cref{sec:supp-cost}).
\item Simulator validation: the FLIP side of V2 was repeated at the evaluation settings (above).
\end{itemize}

\end{document}